\documentclass{llncs}
\usepackage{mathptmx}       
\usepackage{helvet}         
\usepackage{courier}        
\usepackage{type1cm}        
\usepackage{makeidx}         
\usepackage{graphicx}        
\usepackage{multicol}        
\usepackage[bottom]{footmisc}
\usepackage{subfigure}
\usepackage{amsfonts}
\usepackage[cmex10]{amsmath}

\usepackage{xcolor}%

\begin{document}

\title{On the definition of toxicity in NLP}
\titlerunning{Running title}  
%
\author{Sergey Berezin, Reza Farahbakhsh,
Noel Crespi}
\authorrunning{Sergey Berezin et al.} 

\institute{SAMOVAR, Télécom SudParis, Institut Polytechnique de Paris   \\
 91120 Palaiseau, France\\
\email{sberezin@telecom-sudparis.eu}}

\maketitle    
\textit{Abstract.} The fundamental problem in toxicity detection task lies in the fact that the toxicity is
ill-defined. This causes us to rely on subjective and vague data in models' training, which results in non-robust and non-accurate results: garbage in - garbage out.  
This work suggests a new, stress-level-based definition of toxicity designed to be objective and
context-aware. On par with it, we also describe possible ways of applying this new definition to dataset creation and model training.

\section{Introduction}

The toxicity detection task's fundamental problem lies in the toxicity being ill-defined. \\
Jigsaw, a unit within Google and one of the leaders in the field uses a definition of toxicity given by Dixon et al. \cite{def_old} -
"rude, disrespectful, or unreasonable language that is likely to make someone leave a discussion". One can instantly see the issue with this definition, as it gives no quantitative measure of the toxicity and operates with highly subjective cultural terms. Despite all vagueness and flaws, this definition is widely used by many researchers \cite{survey}. 

More than that, lack of proper definition causes researchers to use different terms for similar work \cite{vidgen-etal-2019-challenges}, thus producing labels “toxic", “hateful", “offensive", “abusive", etc. for effectively the same data \cite{its_all_same}. 

The subjectivity of current definitions also causes disagreement in labelling, causing drastically different results in models' training \cite{authors1} 
. Another problem lies in the assumption that toxicity labelling can be done without a proper understanding of the context of each message \cite{context3}. 

This causes racial, sexual, political, religious and geographical bias in datasets \cite{survey} \cite{bias_nig} and also significantly lowers the performance of toxicity detectors, effectively turning them into profanity detectors \cite{holy_shit} without any contextual awareness \cite{xenos-etal-2021-context}.

\section{In search for a formal definition}

First of all, we should be able to measure toxicity objectively. 

In machine learning, we need to have a target metric, such as an amount of money or a category of an animal, that we approximate with some kind of loss function, which we use to solve the optimisation task. Without a clear and measurable target metric, we are unable to confidently tell if our loss function, and therefore our model, is any good at all.

To find an appropriate metric, first of all, we need to think of the objective we want to achieve. The main objective of toxicity detection is to save people from stress, which is caused by disturbances in social interactions, undesirable social roles, criticism, self-criticism and unfair treatment \cite{soc_stress_biosignals}. The stress-evoking mechanism lies in the fact that insults directed at a person pose a severe threat to them and their reputation. More than that, even witnessing the infliction of harm on others causes stress since it shows potential group aggressors and signals social conflict in the vicinity \cite{struiksma2022people}.

So, the key metric for us is the level of toxicity-caused stress a person experiences. 

However, what and why is considered an insult, criticism or any other undesirable interaction?

Let us follow the process of word understanding. After we hear a word, we step into the lexical retrieval phase, during which words are compared to the known social intention or emotional stance of the speaker using this word, the impact on listeners of this word, or the nature of the things referred to by this word.
After understanding these correlations, the release of stored emotional meaning is used as a prediction of what the word means in this particular context \cite{snefjella2020emotion}.

So, our reaction to words is based on our knowledge of the meaning of the words and their acceptance in our morality and communication norms \cite{cialdini2004social}. We, as a society, define norms to prevent what causes stress to us, and we, as a person, feel stress when those norms are violated  \cite{van2019dynamic} \cite{sapolsky2004zebras}.
From that, we can see that toxicity causes stress by contradiction of accepted morality and norms of communication. 

However, morality and norms of communication vary a lot across the world and even across the time \cite{eriksson2021perceptions}. \cite{van2019dynamic}.
While we should not ban a British person for saying that they are going to have a “fag break" when they are going to smoke a cigarette, we should definitely ban somebody who is using the word “fag" as a derogatory name for a homosexual person. In the same way, the word “nice" is perfectly fine in modern English but should be considered toxic when analysing archive letters from the 14th century since it meant “foolish" back then.
This leads us to include situational and verbal context in our account.

Using that, let us formulate the definition: \textbf{Toxicity} - is a characteristic of causing stress by contradiction of accepted morality and norms of interaction with respect to the situational and verbal context of interaction.

From this: \textbf{Toxic speech} - is such a speech that causes stress by contradiction of accepted morality and norms of communication with respect to the situational and verbal context of interaction.

\subsection{Measurements}

The question arises: How do we determine the toxicity of something? 

We can do so by measuring the level of stress. The first way to do so is to register increased levels of stress hormones (e.g. cortisol or catecholamine levels), which can be measured in blood, saliva or urine samples. However, this might not be easy to do en masse. 

We can also use other, non-invasive methods, such as registration of changes in heart rate, blood pressure, pupil diameter, breathing pattern, galvanic skin response, emotion, voice intonation, and body pose \cite{sharma2012objective}. It is shown that using a combination of such measurements can achieve very high quality ($>$ 90\% acc.) estimation of stress levels while keeping the testing procedure fast and simple \cite{soc_stress_biosignals}. 


An even more scalable method is to approach the question from the side of the violation of accepted morality and norms. It is shown that those violations cause stress (i.e., they can be used as a proxy in stress registering) and cause negative responses from other people. By noticing these responses, we can detect the fact of norm violation.

More than that, responses are typically intended to modify the violator's behaviour and to strengthen a violated norm - this gives us a way to not only detect the fact of norm violation but to pinpoint a specific norm and the way of its defyition.\cite{van2015social}. 

A negative reaction to norms' breaching is culture-universal and proportional to the inappropriateness of the triggering behaviour. People consider it more appropriate to use gossip, social isolation, and confrontation the more severe violation of norms is perceived to be. Thus, by registering a negative reaction, we can also estimate the harshness of a norm violation and, subsequently, the level of stress this violation causes \cite{gross2022people}. 

Cross-cultural comparisons have also highlighted how idiosyncratic social norms are. What is considered appropriate in one culture can be seen as highly deviant in another culture. However, norms are consistent within countries and largely independent of the domain of a norm violation \cite{eriksson2021perceptions}.



\subsection{Application and further development}
An automatic way of determining social norms and their violations is described in \cite{chandrasekharan2018internet}. In this work, authors used language models to analyse 2.8M comments removed by moderators of 100 top Reddit sub-forums over ten months. 

Following their idea, we suggest use LLM for creating a dataset based on the proposed definition. The steps are as follows:
\begin{enumerate}
\item Fine-tune a model such as Llama 2 or GPT-4 on corpora of fixed context (i.e. time, region, social group, etc.). An example of such might be a corpus of British literature of the 17th century.
\item Use this fine-tuned model to detect norms' breaches by negative reactions caused by them. Few-shot learning with hand-crafted examples from the corpora might be applied here to ease the task. 
\item Extract the norms violated and label them.
\end{enumerate}
%
This will allow us to build situational and verbal context-aware models for toxic speech detection and will help us solve existing drawbacks of toxicity detection, eliminating its biases and increasing its performance. 


\bibliographystyle{unsrt}
\bibliography{refs} 

\end{document}